\documentclass[10pt,twocolumn,letterpaper]{article}

\usepackage{iccv}
\usepackage{times}
\usepackage{epsfig}
\usepackage{graphicx}
\usepackage{amsmath}
\usepackage{amssymb}
\usepackage{booktabs}
\usepackage{caption}
\usepackage{subcaption}
\usepackage{multirow}
\usepackage{color, colortbl}
\usepackage{fancyvrb}

\DeclareMathOperator*{\argmax}{arg\,max}

\definecolor{Gray}{gray}{0.9}

\usepackage[pagebackref=true,breaklinks=true,letterpaper=true,colorlinks,bookmarks=false]{hyperref}

\iccvfinalcopy 

\def\iccvPaperID{851} 

\ificcvfinal\fi
\begin{document}

\title{Spatial-Aware Object Embeddings for Zero-Shot Localization\\and Classification of Actions}

\author{Pascal Mettes and Cees G. M. Snoek\\
University of Amsterdam
}

\maketitle

\begin{abstract}
We aim for zero-shot localization and classification of human actions in video. Where traditional approaches rely on global attribute or object classification scores for their zero-shot knowledge transfer, our main contribution is a spatial-aware object embedding. To arrive at spatial awareness, we build our embedding on top of freely available actor and object detectors. Relevance of objects is determined in a word embedding space and further enforced with estimated spatial preferences. Besides local object awareness, we also embed global object awareness into our embedding to maximize actor and object interaction. Finally, we exploit the object positions and sizes in the spatial-aware embedding to demonstrate a new spatio-temporal action retrieval scenario with composite queries. Action localization and classification experiments on four contemporary action video datasets support our proposal. Apart from state-of-the-art results in the zero-shot localization and classification settings, our spatial-aware embedding is even competitive with recent supervised action localization alternatives.
\end{abstract}

\section{Introduction}
\begin{figure}[t]
\centering
\includegraphics[width=0.9\linewidth]{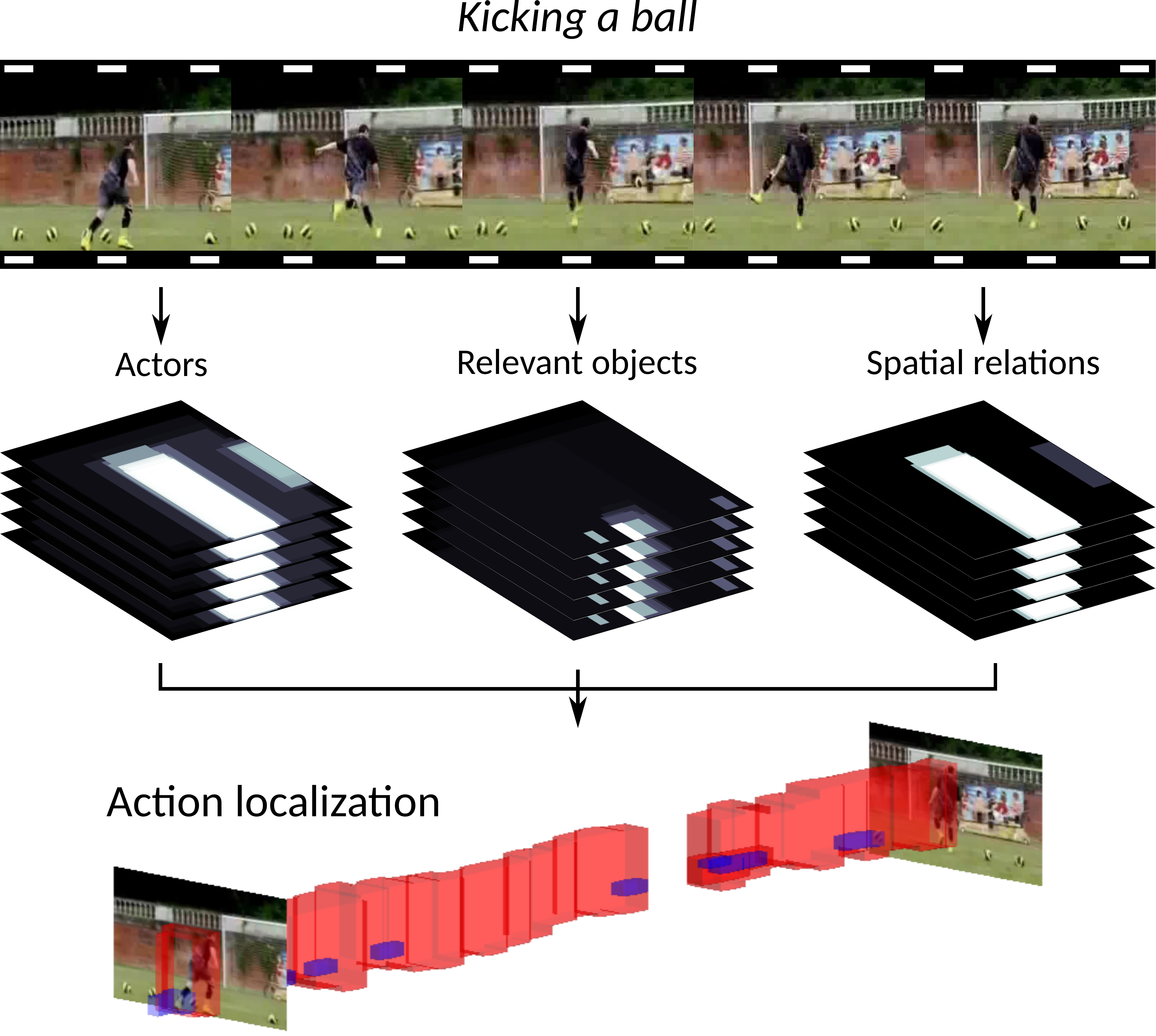}
\caption{\textbf{Spatial-aware object embedding.} Actions are localized and classified by information about actors, relevant objects, and their spatial relations.}
\label{fig:fig1}
\end{figure}

%
We strive for the localization and classification of human actions like \emph{Walking a dog} and \emph{Skateboarding} without the need for any video training examples.
The common approach in this challenging zero-shot setting is to transfer action knowledge via a semantic embedding build from attributes \cite{kodirov2015unsupervised,liu2011recognizing,xu2016multi} or objects \cite{cappallo2016video,jain2015objects2action,wu2016cvpr}. As the semantic embeddings are defined by image or video classifiers, they are unable, nor intended, to capture the spatial interactions an actor has with its environment.  Hence, it is hard to distinguish who is \emph{Throwing a baseball} and who is \emph{Hitting a baseball} when both actions occur within the same video.
We propose a spatial-aware object embedding for localization and classification of human actions in video, see Figure~\ref{fig:fig1}.



We draw inspiration from the \textit{supervised} action classification literature, where the spatial connection between actors and objects has been well recognized, \eg~\cite{gupta2007actionobject,mettes2017localizing,moore1999context,wu2007cvpr}. Early work focused on capturing actors and objects implicitly in a low-level descriptor~\cite{choi20action,yao2010grouplet}, while more recently the benefit of explicitly representing detected objects~\cite{Felzenszwalb10}, their scores, and spatial properties was proven effective~\cite{hanICCV09selectionAndContext,non-localcues_bmvc10,yao2010mutual,yao2011mutual}. Both~\cite{Escorcia2013iccvw} and~\cite{Prest13} demonstrate the benefit of temporal actor and object interaction, by linking detected bounding boxes over time via trackers. By doing so, they are also capable of (supervised) action localization. We also detect actors and objects, and link them over time to capture spatio-temporal interactions. Different from all of the above works, we do no rely on any action class and/or action video supervision to get to our recognition.
Instead, we introduce an embedding built upon actor and object detectors that allows for zero-shot action classification and localization in video.


Our main contribution is a spatial-aware object embedding for zero-shot action localization and classification.
The spatial-aware embedding incorporates word embeddings, box locations for actors and objects, as well as their spatial relations, to generate action tubes.
This enables us to both classify videos and to precisely localize where actions occur.
Our spatial-aware embedding is naturally extended with contextual awareness from global objects.
We furthermore show how our embedding generalizes to any query involving objects, spatial relations, and their sizes in a new spatio-temporal action retrieval scenario.
Action localization and classification experiments on four contemporary action video datasets support our proposal.

\section{Related work}

\subsection{Supervised action localization and classification}
A wide range of works have proposed representations to classify actions given video examples. Such representations include local spatio-temporal interest points and features~\cite{laptev2008learning,liu2016learning,willems2008efficient} and local trajectories~\cite{amor2016action,wang2013action}, typically aggregated into VLAD or Fisher vector representations~\cite{oneata2013action,peng2014action}. Recent works focus on learning global representations from deep networks, pre-trained on optical flow~\cite{simonyan2014two} or large-scale object annotations~\cite{jain2015what,karpathy2014large,xu2015discriminative}.
We also rely on deep representations for our global objects, but we emphasize on local objects and we aim to classify and localize actions without the need for any video example.

For spatio-temporal action localization, a popular approach is to split videos into action proposals; spatio-temporal tubes in videos likely to contain an action. Annotated tubes from example videos are required to train a model to select the best action proposals at test time. Action proposal methods include merging supervoxels~\cite{jain2014action,soomro2015action}, merging trajectories~\cite{chen2015action,mettes2016spot}, and detecting actors~\cite{yu2015fast}. The current state-of-the-art action localizers employ Faster R-CNN~\cite{ren2015faster} trained on bounding box annotations of actions in video frames~\cite{gkioxari2015finding,weinzaepfel2015learning}. We are inspired by the effectiveness of actor detections and Faster R-CNN for localization, but we prefer commonly available detectors trained on images. We employ these detectors as input to our spatial-aware embedding for localization in video in a zero-shot setting.

\subsection{Zero-shot action localization and classification}

Inspired by zero-shot image classification~\cite{lampert2014attribute}, several works have performed zero-shot action classification by learning a mapping of actions to attributes~\cite{gan2016learning,liu2011recognizing,zhang2015robust}. Models are trained for the attributes from training videos of other actions and used to compare test videos to unseen actions. Attribute-based classification has been extended \eg using transductive learning~\cite{fu2014transductive,xu2017transductive} and domain adaption~\cite{kodirov2015unsupervised,xu2016multi}. Due to the necessity to manually map each action to global attributes a priori, these approaches do not generalize to arbitrary zero-shot queries and are unable to localize actions, which is why we do not employ attributes in our work.

Rather than mapping actions to attributes, test actions can also be mapped directly to actions used for training.
Li \etal~\cite{li2016recognizing} map visual video features to a semantic space shared by training and test actions.
Gan \etal~\cite{gan2016recognizing} train a classifier for an unseen action by relating the action to training actions at several levels of relatedness.
Although the need for attributes is relieved with such mappings, this approach still requires videos of other actions for training and is only able to classify actions. We localize and classify actions without using any videos of actions during training.

A number of works have proposed zero-shot classification by exploiting large amounts of image and object labels~\cite{imagenet09}.
Given deep networks trained on image data, these approaches map object scores in videos to actions \eg using word vectors~\cite{cappallo2016video,inoue2016adaptation,jain2015objects2action,wu2016cvpr} or auxiliary textual descriptions~\cite{gan2016concepts,habibian2016videostory,wu2014cvpr}.
Objects as the basis for actions results in effective zero-shot classification and generalizes to arbitrary actions.
However, these approaches are holistic; object scores are computed over whole videos.
In this work, we take the object-based perspective to a local level, which allows us to model the spatial interaction between actors and objects for action localization, classification, and retrieval.

The work of Jain \etal~\cite{jain2015objects2action} has previously performed zero-shot action localization. Their approach first generates action proposals. In a second pass, each proposal is represented with object classification scores. The proposals best matching the action name in word2vec space are selected. Their approach does not use any object detectors, nor is there any explicit notion of spatial-awarenes inside each action proposal. Finally, spatial relations between actors and objects are ignored. As we will show in the experiments, inclusion of our spatial-awareness solves these limitations and leads to a better zero-shot action localization.

\section{Spatial-aware object embeddings}
In our zero-shot formulation, we are given a set of test videos $\mathcal{V}$ and a set of action class names $\mathcal{Z}$. We aim to classify each video to its correct class and to discover the spatio-temporal tubes encapsulating each action in all videos. To that end, we propose a spatial-aware embedding; scored action tubes from interactions between actors and local objects. We present our embeddings in three steps: (i) gathering prior knowledge on actions, actors, objects, and their interactions, (ii) computing spatial-aware embedding scores for bounding boxes, and (iii) linking boxes into action tubes.

\begin{figure}[t]
\centering
\begin{subfigure}{0.31\linewidth}
\includegraphics[width=\linewidth]{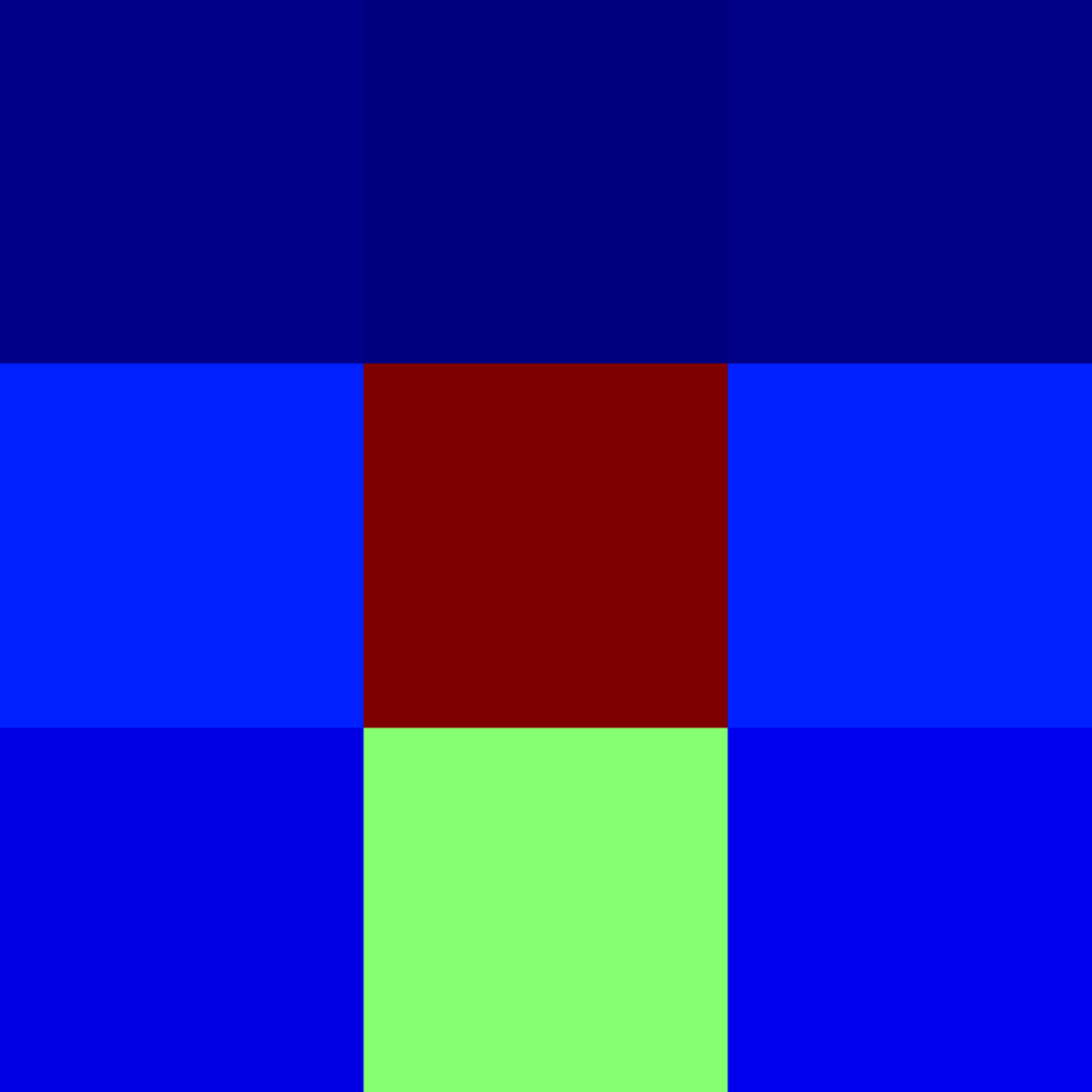}
\caption{Skateboard.}
\label{fig:grid-b}
\end{subfigure}
\begin{subfigure}{0.31\linewidth}
\includegraphics[width=\linewidth]{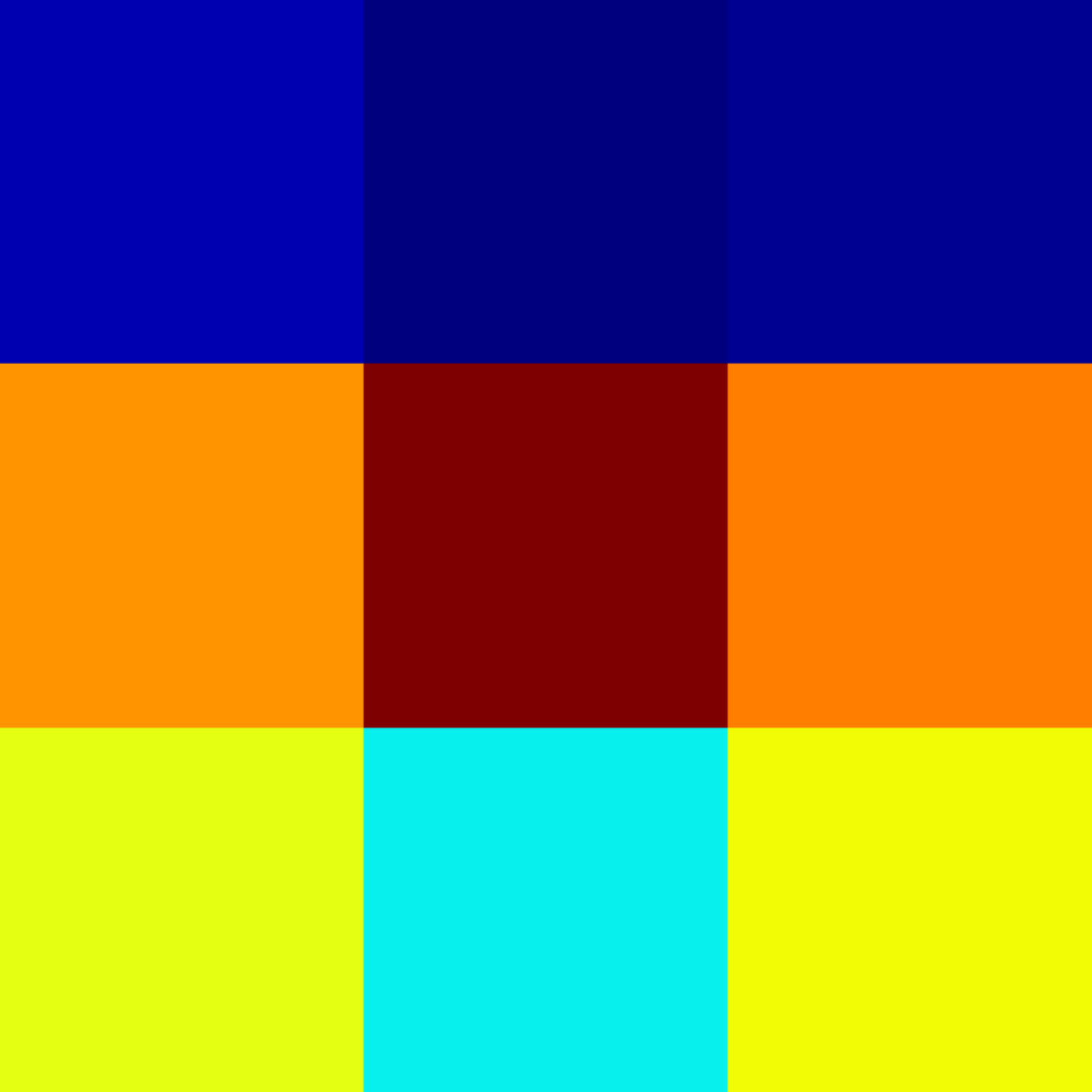}
\caption{Bicycle.}
\label{fig:grid-c}
\end{subfigure}
\begin{subfigure}{0.31\linewidth}
\includegraphics[width=\linewidth]{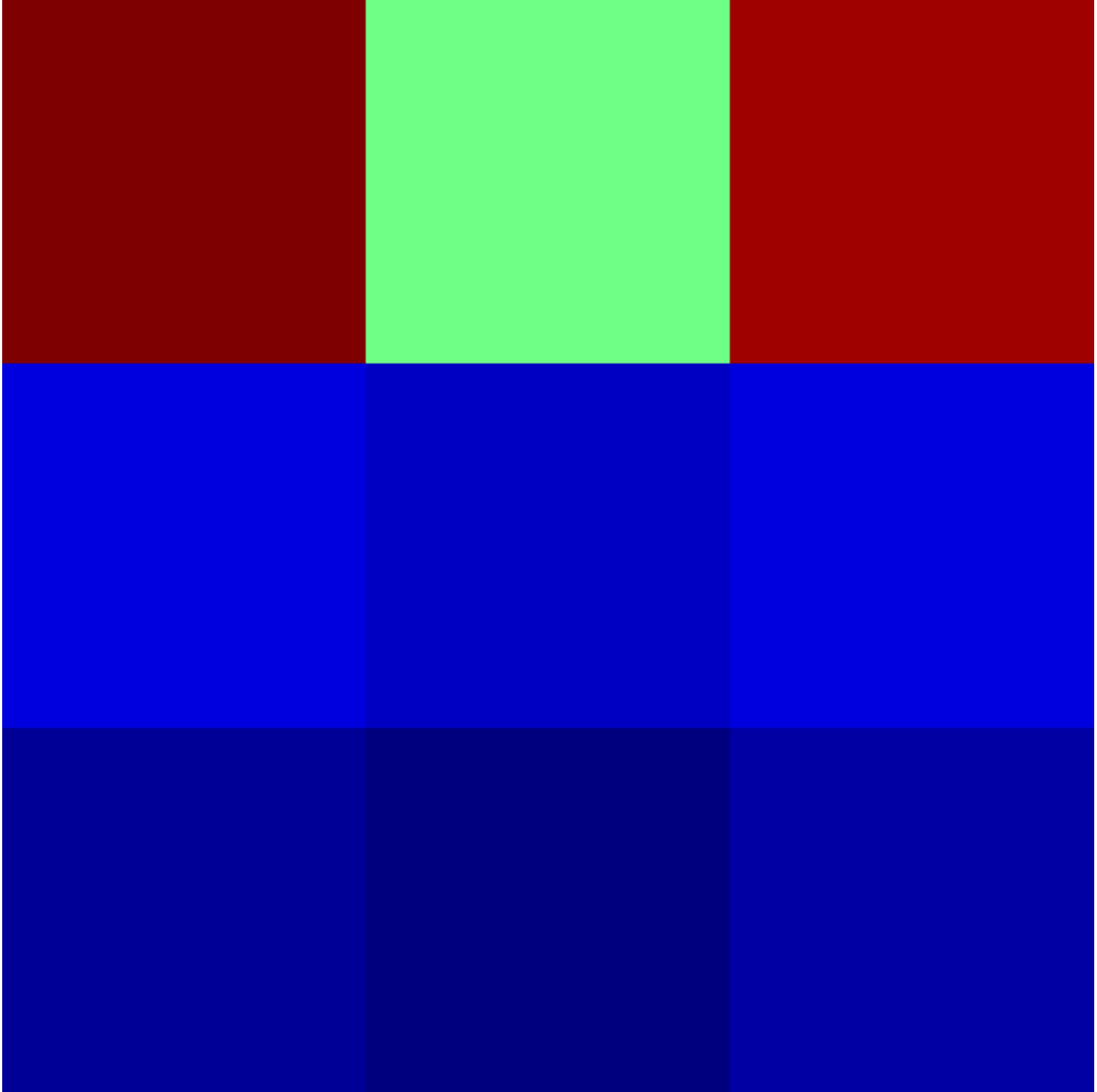}
\caption{Traffic light.}
\label{fig:grid-d}
\end{subfigure}
min \includegraphics[width=0.6\linewidth]{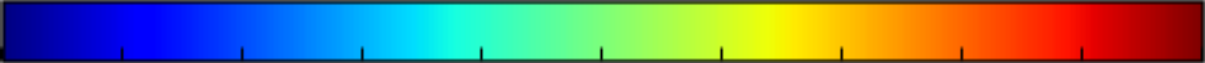} max
\caption{\textbf{Examples of preferred spatial relations of objects relative to actors.} In line with our intuition, skateboards are typically on or below the actor, while bicycles are typically to the left or right of actors and traffic lights are above the actors.}
%
%
\label{fig:grid}
\end{figure}

\subsection{Prior knowledge}
\textbf{Local object detectors.} We first gather a set of local detectors pre-trained on images. Let $\mathcal{O} = \{O_{D}, O_{N}\}$ denote the objects with detectors $O_{D}$ and names $O_{N}$. Furthermore, let $\mathcal{A} = \{A_{D}, \texttt{actor}\}$ denote the actor detector. Each detector outputs a set of bounding boxes with corresponding object probability scores per video frame.

\textbf{Textual embedding.} Given an action class name $Z \in \mathcal{Z}$, we aim to select a sparse subset of objects $\mathcal{O}_{Z} \subset \mathcal{O}$ relevant for the action. For the selection, we rely on semantic textual representations as provided by word2vec~\cite{mikolov2013distributed}. The similarity between object $o$ and the action class name is given as:
\begin{equation}
w(o, Z) = \text{cos}(e(o_{N}), e(Z)),
\end{equation}
where $e(\cdot)$ states the word2vec representation of the name. We select the objects with maximum similarity to the action.

%
\textbf{Actor-object relations.} We exploit that actors interact with objects in preferred spatial relations. To do so, we explore where objects tend to occur relative to the actor. Since we can not learn precise spatial relations between actors and objects from examples, we aim to use common spatial relations between actors and objects, as can be mined from large-scale image data sets. We discretize the spatial relations into nine relative positions, representing the preposition \emph{in front of} and the eight basic prepositions around the actor, \ie \emph{left of, right of, above, below}, and the four corners (\eg \emph{above left}). For each object, we obtain a nine-dimensional distribution specifying its expected location relative to the actor, as detailed in Figure~\ref{fig:grid}.


\begin{figure*}[t]
\centering
\begin{subfigure}{0.245\textwidth}
\includegraphics[width=\textwidth]{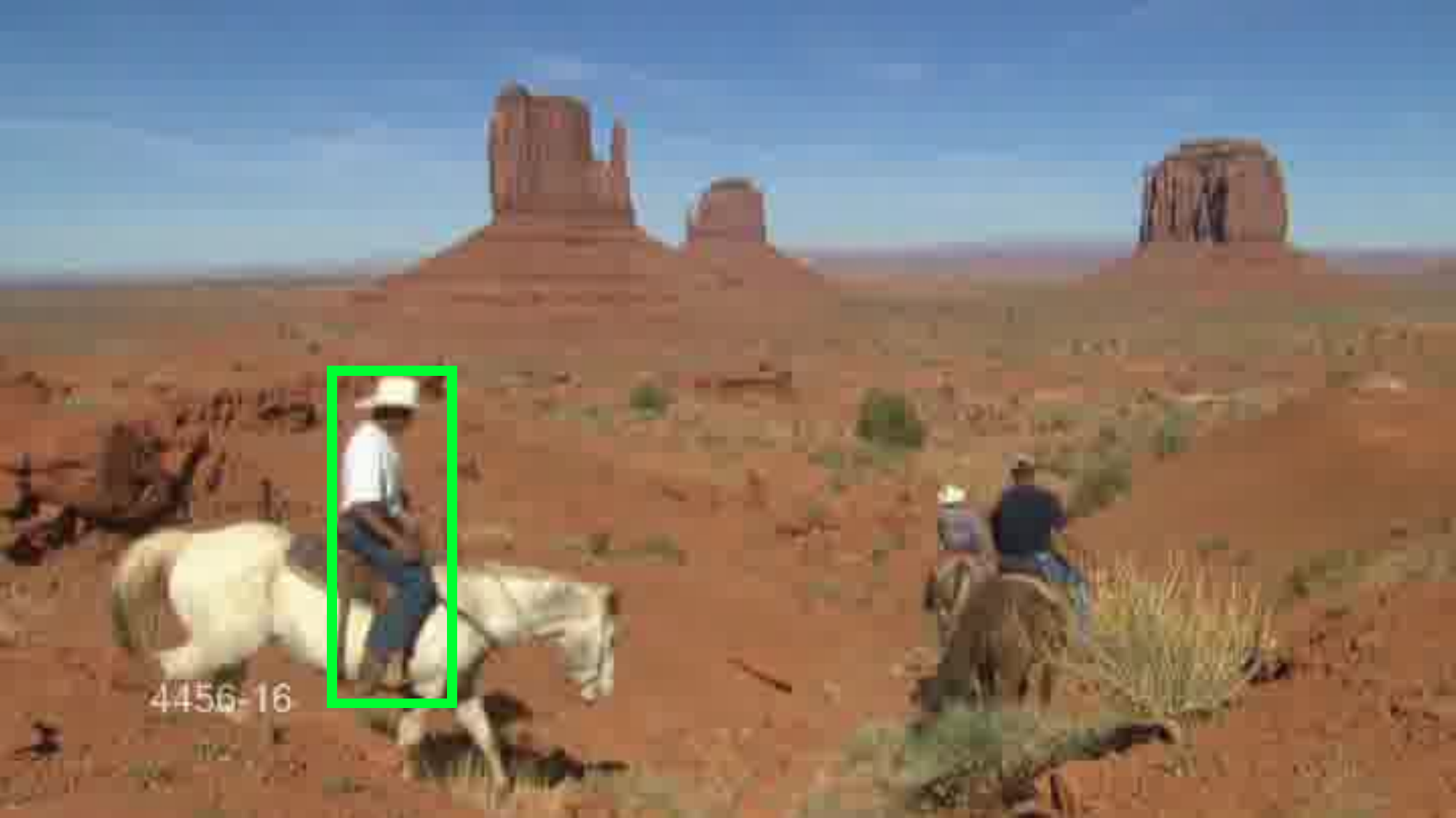}
\caption{Video frame.}
\label{fig:actorboxes-a}
\end{subfigure}
\begin{subfigure}{0.245\textwidth}
\includegraphics[width=\textwidth]{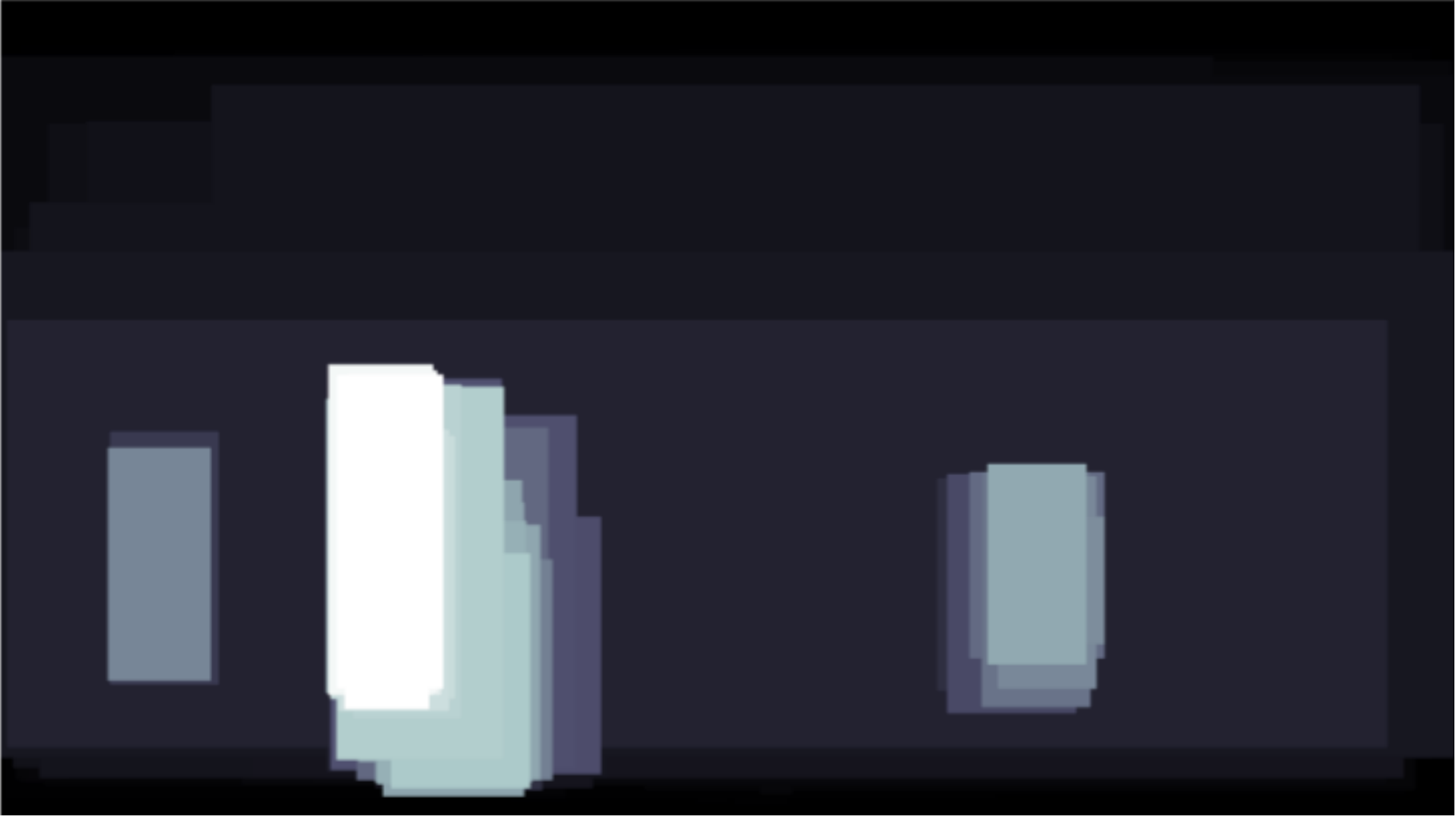}
\caption{Actor detection.}
\label{fig:actorboxes-b}
\end{subfigure}
\begin{subfigure}{0.245\textwidth}
\includegraphics[width=\textwidth]{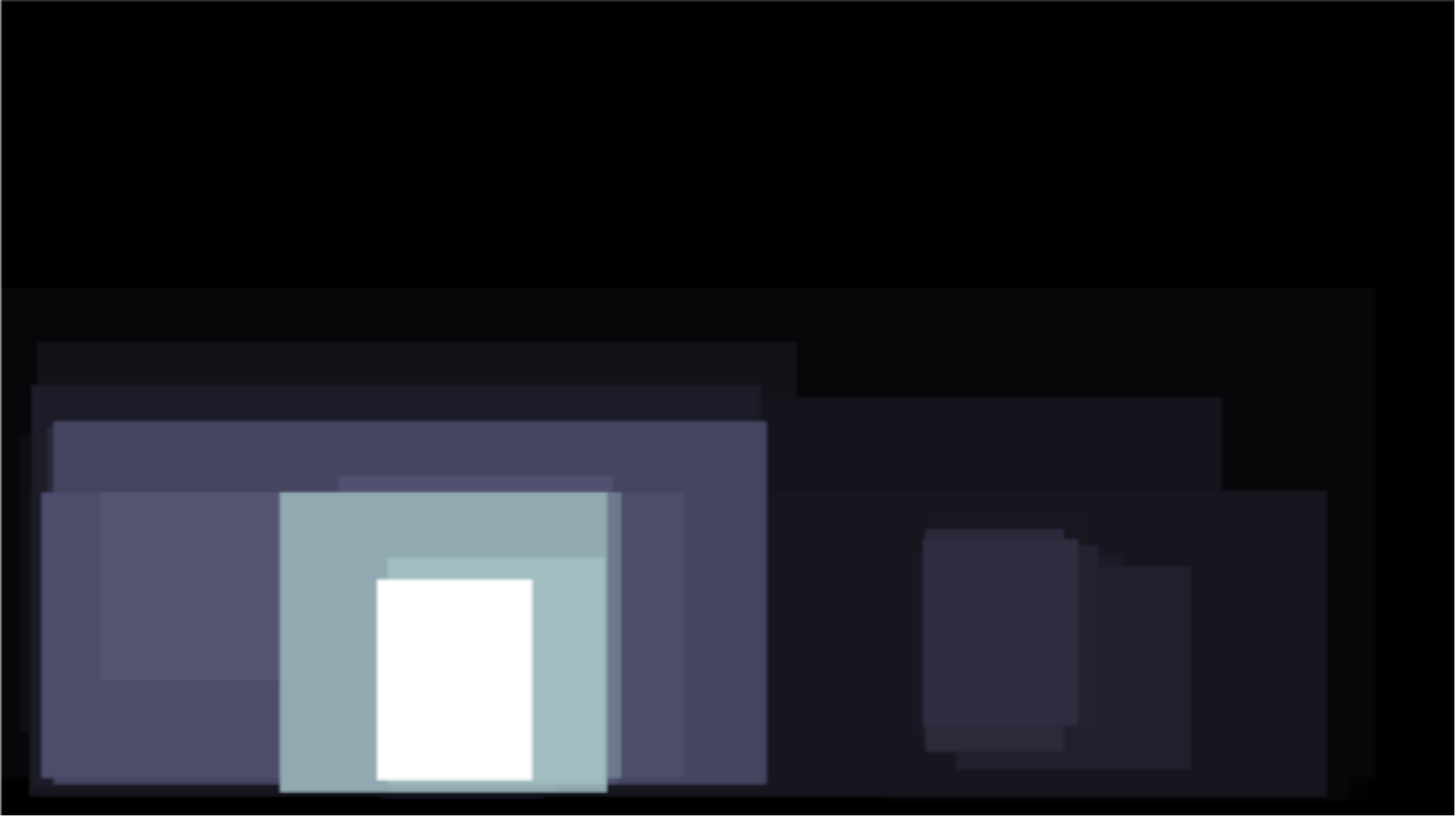}
\caption{Object detection (horse).}
\label{fig:actorboxes-c}
\end{subfigure}
\begin{subfigure}{0.245\textwidth}
\includegraphics[width=\textwidth]{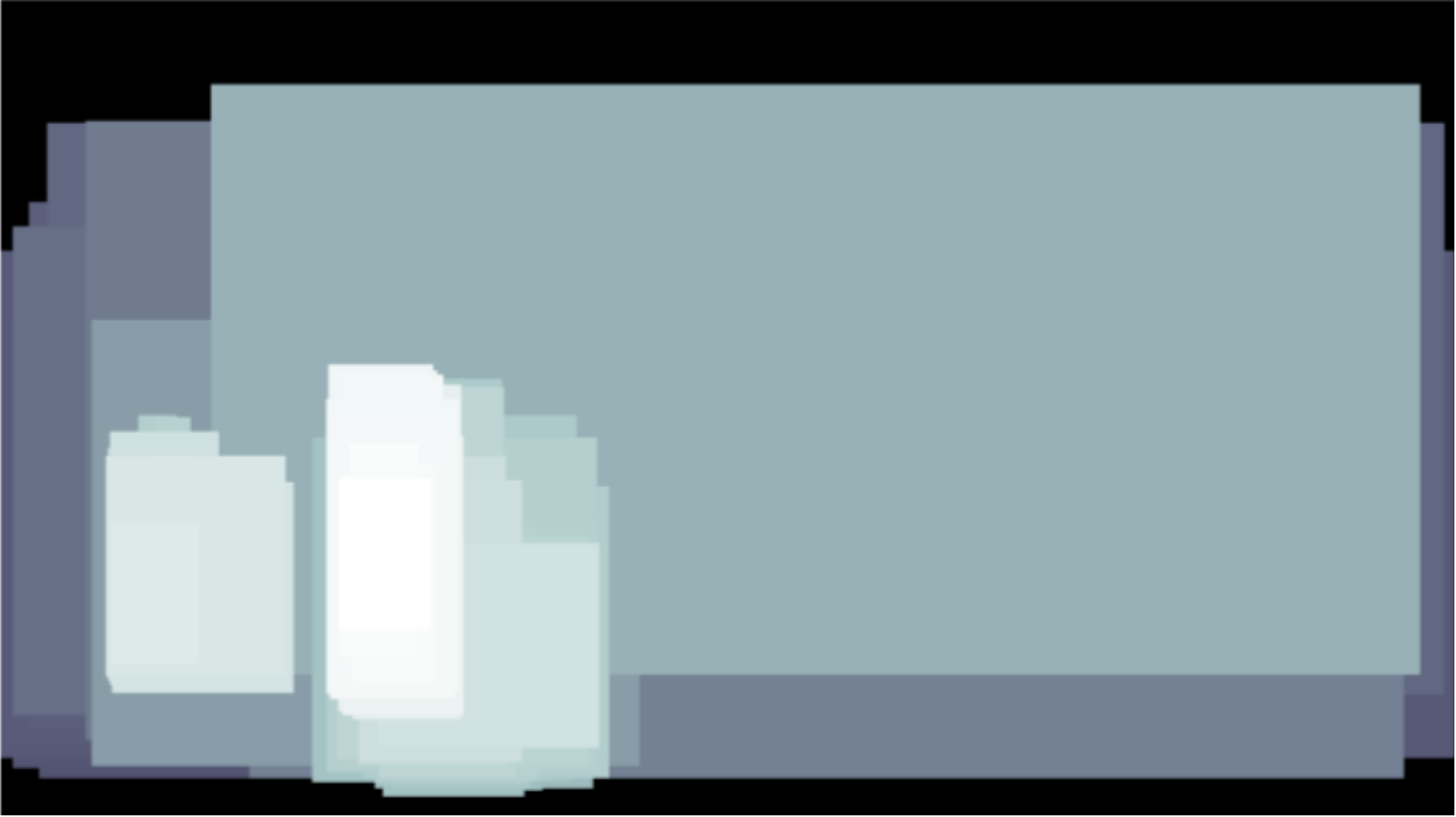}
\caption{Spatial relation match.}
\label{fig:actorboxes-d}
\end{subfigure}
\caption{\textbf{Example of our spatial-aware embedding.} The actor sitting on the left horse (green box) is most relevant for the action \emph{Riding horse} based on the actor detection, horse detection, and spatial relations between actors and horses.}
\label{fig:actorboxes}
\end{figure*}

\subsection{Scoring actor boxes with object interaction}
We exploit our sources of prior knowledge to compute a score for the detected bounding boxes in all frames of each test video $V \in \mathcal{V}$.
Given a bounding box $b$ in frame $F$ of video $V$, we define a score function that incorporates the presence of (i) actors, (ii) relevant local objects, and (iii) the preferred spatial relation between actors and objects. A visual overview of the three components is shown in Figure~\ref{fig:actorboxes}.
More formally, we define a score function for box $b$ given an action class $Z$ as:
\begin{equation}
s(b, F, Z) = p(A_D | b) + \sum_{o \in \mathcal{O}_{\mathcal{Z}}} r(o, b, F, Z),
\label{eq:abp}
\end{equation}
where $p(A_D | b)$ is the probability of an actor being present in bounding box $b$ as specified by the detector $A_D$. The function $r$ expresses the object presence and relation to the actor, it is defined as:
\begin{equation}
r(o, b, F, Z) = w(o, Z) \cdot \bigg( \max_{f \in F_{n}} p(o_D | f) \cdot m(o, \mathcal{A}, b, f) \bigg),
\label{eq:object-match}
\end{equation}
where $w(o, Z)$ states the semantic relation score between object $o$ and action $Z$ and $F_{n}$ states all bounding boxes within the neighourhood of box $b$ in frame $F$. The second part of Equation~\ref{eq:object-match} states that we are looking for a box $f$ around $b$ that maximizes the joint probability of the presence of object $o$ (the function $p(o_D | f)$), the match between the spatial relations of $(b,f)$ and the prior relations of the actor and object $o$ (the function $m$). We define the spatial relation match as:
\begin{equation}
m(o, \mathcal{A}, b, f) = 1 - JSD_{2}(d(\mathcal{A}, o) || d(b, f)),
\label{eq:relmatch}
\end{equation}
where $JSD_{2}(\cdot || \cdot) \in [0,1]$ denotes the Jensen-Shannon Divergence with base 2 logarithm~\cite{lin1991divergence}. Intuitively, the Jensen-Shannon Divergence, a symmetrized and bounded variant of the Kullback-Leibler divergence, determines to what extent the two 9-dimensional distributions match. The more similar the distributions, the lower the divergence, hence the need for the inversion as we aim for maximization.
%


\subsection{Linking spatial-aware boxes}
The score function of Equation~\ref{eq:abp} provides a spatial-aware embedding score for each bounding box in each frame of a video. We apply the score function to the boxes of all actor detections in each frame. We form tubes from the individual box scores by linking them over time~\cite{gkioxari2015finding}. We link those boxes over time that by themselves have a high score from our spatial-aware embedding and have a high overlap amongst each other. This maximization problem is solved using dynamic programming with the Viterbi algorithm. Once we have a tube from the optimization, we remove all boxes from that tube and compute the next tube from the remaining boxes.

Let $T$ denote a discovered action tube in a video. The corresponding score is given as:
\begin{equation}
t_{\text{emb}}(T, Z) = \frac{1}{|T|} \sum_{t \in T} s(t_{b}, t_{F}, Z),
\label{eq:temb}
\end{equation}
where $t_b$ and $t_F$ denote a bounding box and the corresponding frame in tube $T$.

In summary, we propose spatial-aware object embeddings for actions; tubes through videos by linking boxes based on the zero-shot likelihood from the presence of actors, the presence of relevant objects around the actors, and the expected spatial relations between objects and actors.

\section{Local and global object interaction}
%
To distinguish tubes from different videos in a collection, contextual awareness in the form of relevant global object classifiers is also a viable source of information.
Here, we first outline how to obtain video-level scores based on object classifiers.
Then, we show how to compute spatial- and global-aware embeddings for action localization, classification, and retrieval.

\subsection{Scoring videos with global objects}
Let $\mathcal{G} = \{G_{C}, G_{N}\}$ denote the set of global objects with corresponding classifiers and names. Different from the local objects $\mathcal{O}$, these objects provide classifier scores over a whole video. Given an action class name $Z$, we again select the top relevant objects $\mathcal{G}_{Z} \subset \mathcal{G}$ using the textual embedding.
The score of a video $V$ is then computed as a linear combination of the word2vec similarity and classifier probabilities over the top relevant objects:
\begin{equation}
t_{\text{global}}(V, Z) = \sum_{g \in G_{Z}} w(g, Z) \cdot p(g | V),
\label{eq:tglobal}
\end{equation}
where $p(g | V)$ denotes the probability of global object $g$ of being in video $V$.

\subsection{Spatial- and global-aware embedding}
The information from local and global objects is combined into a spatial- and global-aware embedding. Here, we show how this embedding is employed for spatial-aware action localization, classification, and retrieval.

\textbf{Action localization.}
For localization, we combine the tube score from our spatial-aware embedding with the video score from the global objects into a score for each individual tube $T$ as:
\begin{equation}
t(T, V, Z) = t_{\text{emb}}(T, Z) + t_{\text{global}}(V, Z).
\label{eq:tcombined}
\end{equation}
We note that incorporating scores from global objects does not distinguish tubes from the same video. The global scores are however discriminative for distinguishing tubes from different videos in a collection $\mathcal{V}$. We compute the final score for all tubes of all videos in $\mathcal{V}$ using Equation~\ref{eq:tcombined}. We then select the top scoring tubes per video, and rank the tubes over all videos based on their scores for localization.

\begin{table*}[t]
	\centering
		\scalebox{0.99}{
			\begin{tabular}{l r r r r r r r r}
				\toprule
				& \multicolumn{4}{c}{\textbf{Localization} (mAP @ 0.5)} & \multicolumn{4}{c}{\textbf{Classification} (mean accuracy)}\\
				\cmidrule(lr){2-5} \cmidrule(lr){6-9}
				& \multicolumn{4}{c}{\# local objects} & \multicolumn{4}{c}{\# local objects}\\
				 & 0 & 1 & 2 & 5 & 0 & 1 & 2 & 5\\
				\midrule
				Embedding I:  \textbf{\textit{Actor-only}} & 0.083 & - & - & - & 0.100 & - & - & -\\
				Embedding II: \textbf{\textit{Actors and objects}} 	& - & 0.175 & 0.182 & 0.193 & - & 0.205 & 0.117 & 0.139\\
				Embedding III: \textbf{\textit{Spatial-aware}} & - & \textbf{0.221} & 0.209 & 0.199 & - & 0.180 & 0.196 & \textbf{0.255}\\
				\bottomrule
			\end{tabular}
		}
	\caption{\textbf{Influence of spatial awareness.} On UCF Sports we compare our spatial-aware object embedding to two other embeddings; using only the actors and using actors with objects, while ignoring their spatial relations. Our spatial-aware embedding is preferred for both localization (one object per action) and classification (five objects per action).}
	\label{tab:exp1}
\end{table*}

\textbf{Action classification.}
For classification purposes, we are no longer concerned about the precise location of the tubes from the spatial-aware embeddings. Therefore, we compute the score of a video $V$ given an action class name $Z$ using a max-pooling operation over the scores from all tubes $T_{V}$ in the video. The max-pooled score is then combined with the video score from the global objects. The predicted class for video $V$ is determined as the class with the highest combined score:
\begin{equation}
c_{V}^{*} = \argmax_{Z \in \mathcal{Z}} \bigg( \max_{T \in T_{V}} t_{\text{emb}}(T, Z) + t_{\text{global}}(V, Z) \bigg).
\label{eq:class}
\end{equation}

\textbf{Spatial-aware action retrieval.}
Spatial-aware action retrieval from user queries resembles action localization, \ie rank the most relevant tubes the highest. However,  different from localization, we now have the opportunity to specify actor and object relations via the search query. Given the effectiveness of size in actor-object interactions~\cite{Escorcia2013iccvw}, we can also allow users to specify a relative object size $r$.
By altering the size of queries objects, different localizations can be retrieved of the same action.
To facilitate spatial-aware action retrieval, we alter the spatial relation match of Equation~\ref{eq:relmatch} with a match for a specified relative object size:
\begin{equation}
q(o, \mathcal{A}, b, f, r) = m(o, \mathcal{A}, b, f) + \bigg(1 - | \frac{s(b)}{s(f)} - r |\bigg),
\label{eq:sizematch}
\end{equation}
where $s(\cdot)$ denotes the size of a bounding box. Substituting the spatial relation match with Equation~\ref{eq:sizematch}, we again rank top scoring tubes, but now by maximizing a match to user-specified objects, spatial relations, and relative size.

\section{Experimental setup}

\subsection{Datasets}


\textbf{UCF Sports} consists of 150 videos from 10 sport action categories, such as \emph{Skateboarding}, \emph{Horse riding}, and \emph{Walking}~\cite{rodriguez2008action}. We employ the test split as suggested in~\cite{lan2011discriminative}.

\textbf{UCF 101} consists of 13,320 videos from 101 action categories, such as \emph{Skiing}, \emph{Basketball dunk}, and \emph{Surfing}~\cite{soomro2012ucf101}. We use this dataset for classification and use the test splits as provided in~\cite{soomro2012ucf101}, unless stated otherwise.

\textbf{J-HMDB} consists of 928 videos from 21 actions, such as~\emph{Sitting}, \emph{Laughing}, and \emph{Dribbling}~\cite{jhuang2013towards}. We use the bounding box around the binary action masks as the spatio-temporal annotations for localization. We use the test split as suggested in~\cite{jhuang2013towards}.

\textbf{Hollywood2Tubes} consists of 1,707 videos from the Hollywood2 dataset~\cite{marszalek09}, supplemented with spatio-temporal annotations for localization~\cite{mettes2016spot}. Actions include \emph{Fighting with a person}, \emph{Eating}, and \emph{Getting out of a car}. We use the test split as suggested in~\cite{marszalek09}.

\subsection{Implementation details}


\textbf{Textual embedding.}
To map the semantics of actions to objects, we employ the skip-gram network of word2vec~\cite{mikolov2013distributed} trained on the metadata of the images and videos from the YFCC100M dataset~\cite{thomee2016yfcc100m}. This model outputs a 500-dimensional representation for each word. If an action or object consists of multiple words, we average the representations of the individual words~\cite{jain2015objects2action}. 
%

\textbf{Actor and object detection.} 
For the detection of both the actors and the local objects, we use Faster R-CNN~\cite{ren2015faster}, pre-trained on the MS-COCO dataset~\cite{lin2014microsoft}. This network consists of the actor class and 79 other objects, such as \emph{snowboard}, \emph{horse}, and \emph{toaster}. After non-maximum suppression, we obtain roughly 50 detections for each object per frame. We apply the network to each frame (UCF Sports, J-HMDB), or each $5^{th}$ frame (UCF 101, Hollywood2Tubes) followed by linear interpolation.

\textbf{Spatial relations.} The spatial relations between actors and objects are also estimated from the MS-COCO dataset. For each object instance, we examine the spatial relations with the closest actor (if any actor is close to the object). We average the relations over all instances for each object.

\textbf{Object classification.}
For the global objects, we employ a GoogLeNet network~\cite{szegedy2015going}, pre-trained on a 12,988-category shuffle~\cite{mettes2016imagenet} of ImageNet~\cite{imagenet09}. This network is applied to each $5^{th}$ frame of each video. For each frame, we obtain the object probabilities at the softmax layer and average the probabilities over the entire video. Following~\cite{jain2015objects2action}, we select the top 100 most relevant objects per action.

\textbf{Evaluation.} For localization, we compute the spatio-temporal intersection-over-union between top ranked actor tubes and ground truth tubes. We report results using both the (mean) Average Precision and AUC metrics. For classification, we evaluate with mean class accuracy.

\section{Experimental results}


\begin{figure*}[t]
\centering
\begin{subfigure}{0.22\textwidth}
\centering
\emph{Skateboarding}\\
\includegraphics[width=\textwidth]{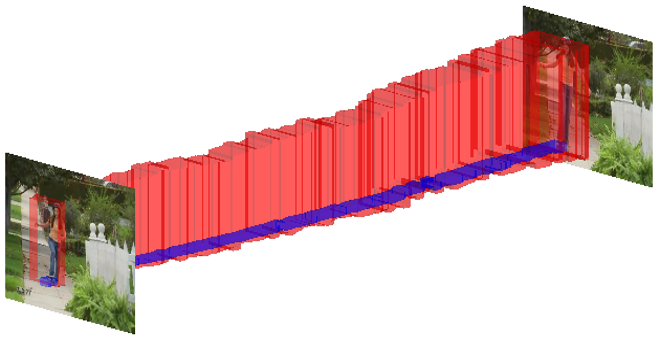}
Top object: \texttt{skateboard}
\end{subfigure}
\hspace{0.5cm}
\begin{subfigure}{0.22\textwidth}
\centering
\emph{Riding a horse}\\
\includegraphics[width=\textwidth]{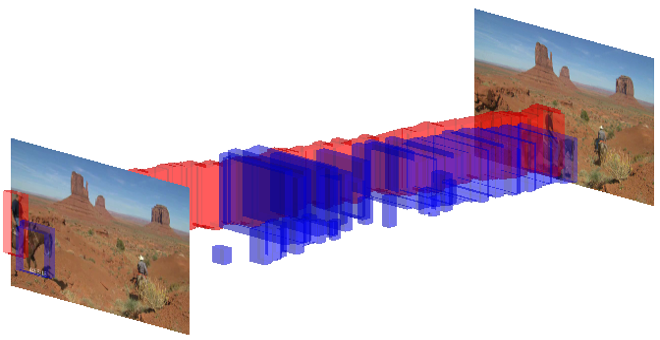}
Top object: \texttt{horse}
\end{subfigure}
\hspace{0.5cm}
\begin{subfigure}{0.22\textwidth}
\centering
\emph{Swinging on a bar}\\
\includegraphics[width=\textwidth]{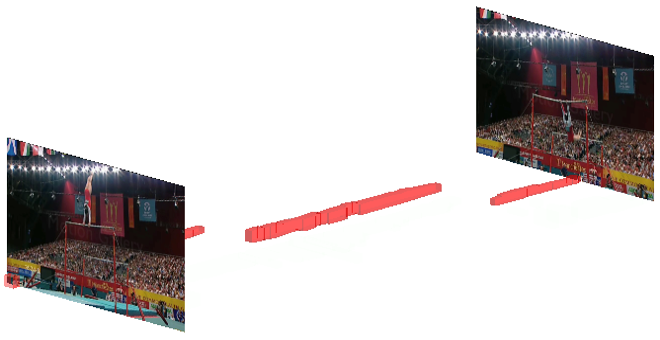}
Top object: \texttt{table}
\end{subfigure}
\hspace{0.5cm}
\begin{subfigure}{0.22\textwidth}
\centering
\emph{Kicking}\\
\includegraphics[width=\textwidth]{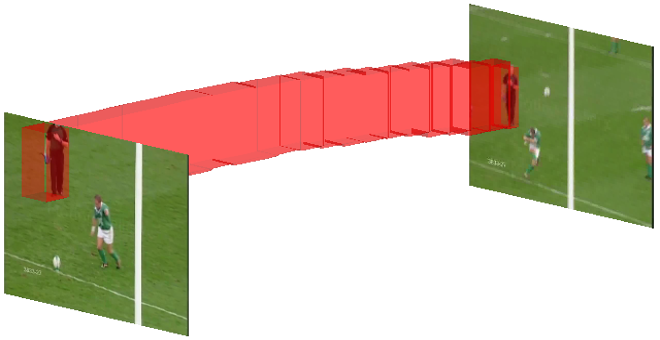}
Top object: \texttt{tie}
\end{subfigure}
\caption{\textbf{Qualitative action localization results.} For \emph{Skateboarding} and \emph{Riding a horse}, relevant objects (blue) aid our localization (red). For \emph{Swinging on a bar} and \emph{Kicking}, incorrectly selected objects result in incorrect localizations. We expect that including more object detectors into our embedding will further improve results.}
\label{fig:exp1-qual}
\end{figure*}

\subsection{Spatial-aware embedding properties}
In the first experiment, we focus on the properties of our spatial-aware embedding, namely the number of local objects to select and the influence of the spatial relations. We also evaluate qualitatively the effect of selecting relevant objects per action. We evaluate these properties on the UCF Sports dataset for both localization and classification.

\textbf{Influence of local objects.}
We evaluate the performance using three settings of our embeddings. The first setting is using solely the actor detections for scoring bounding boxes. The second setting uses both the actor and the top relevant objects(s), but ignores the spatial relations between actors and objects. The third setting is our spatial-aware embedding, which combines the information from actors, objects, and their spatial relations.

In Table~\ref{tab:exp1}, we provide both the localization and classification results.
For localization using only the actor results in tubes that might overlap well with the action of interest, but there is no direct means to separate tubes containing different actions. This results in low Average Precision scores.
For classification, using only the actor results in weak accuracy scores. This is because there is again no mechanism to discriminate videos containing different actions. 

The second row of Table~\ref{tab:exp1} shows the result when incorporating local object detections. For both localization and classification, there is a considerable increase in performance, indicating the importance of detections of relevant objects for zero-shot action localization and classification.



In the third row of Table~\ref{tab:exp1}, we show the performance of our spatial-aware embedding. The embedding outperforms the other settings for both localization and classification. This result shows that gathering and capturing information about the relative spatial locations of objects and actors provides valuable information about actions in videos.
The spatial-aware embedding is most beneficial for the action \emph{Riding a horse} (from 0.03 to 0.75 mAP), due to the consistent co-occurrence of actors and horses.
Contrarily, the performance for \emph{Running} remains unaltered, which is because no object relevant to the action is amongst the available detectors.

We have additionally performed an experiment with finer grid sizes on UCF Sports.
For localization with the top-5 objects, we reach an mAP of 0.170 (4x4 grid) and 0.171 (5x5 grid), compared to a score of 0.199 with the 3x3 grid.
Overall, the scores descrease slightly with finer grid sizes, indicating that coarse spatial relations are preferred over fine spatial relations.

\begin{figure}[t]
\centering
\includegraphics[width=0.9\linewidth]{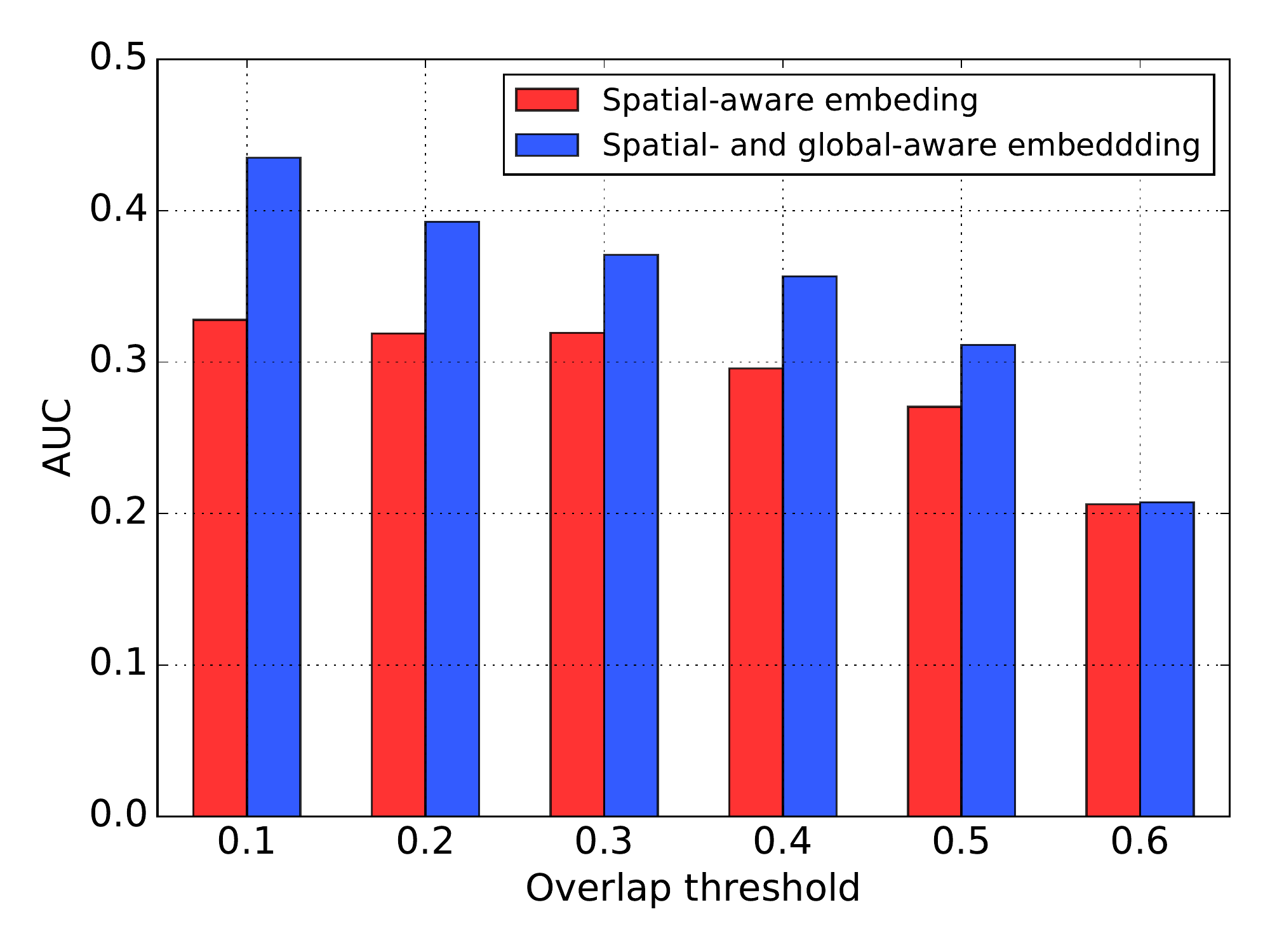}
\caption{\textbf{Local and global object interaction effect on localization.} Adding global object awarereness further improves our spatial-aware object embedding on UCF Sports, especially at low overlap thresholds.}
\label{fig:exp2}
\end{figure}

\textbf{How many local objects?}
In Table~\ref{tab:exp1} we also consider how many relevant local objects to maintain per action. For localization, we observe a peak in performance using the top-1 local object per action, with a mean Average Precision (mAP) of 0.221 at an overlap threshold of 0.5; a sharp increase in performance over the 0.083 mAP using only the actor. When more objects are used, the performance of our embeddings degrades slightly, indicating that actors are more likely to interact with a single object than multiple objects on a local level. At least for the UCF Sports dataset.

For classification, we observe a reverse correlation; the more local objects in our embedding, the higher the classification accuracy. This result indicates that for classification, we want to aggregate more information about object presence in videos, rather than exploit the single most relevant object per action. This is because a precise overlap with the action in each video is no longer required for classification. We exploit this relaxation with the max-pooling operation in the video-level scoring of Equation~\ref{eq:class}.



\textbf{Selecting relevant objects.}
In our zero-shot formulation, a correct action recognition depends on detecting objects relevant to the action.
We highlight the effect of detecting relevant objects in Figure~\ref{fig:exp1-qual}.
For successful actions such as \emph{Skateboarding} and \emph{Riding a horse}, the detection of respectively skateboards and horses help to generate a desirable action localization. For the actions \emph{Swinging on a bar} and \emph{Kicking}, the top selected objects are however incorrect, either because no relevant object is available or because of ambiguity in the word2vec representations.

\begin{figure*}[t]
\centering
\begin{subfigure}{0.3\textwidth}
\centering
\emph{Backpack (0.25) on actor}
\includegraphics[width=\textwidth]{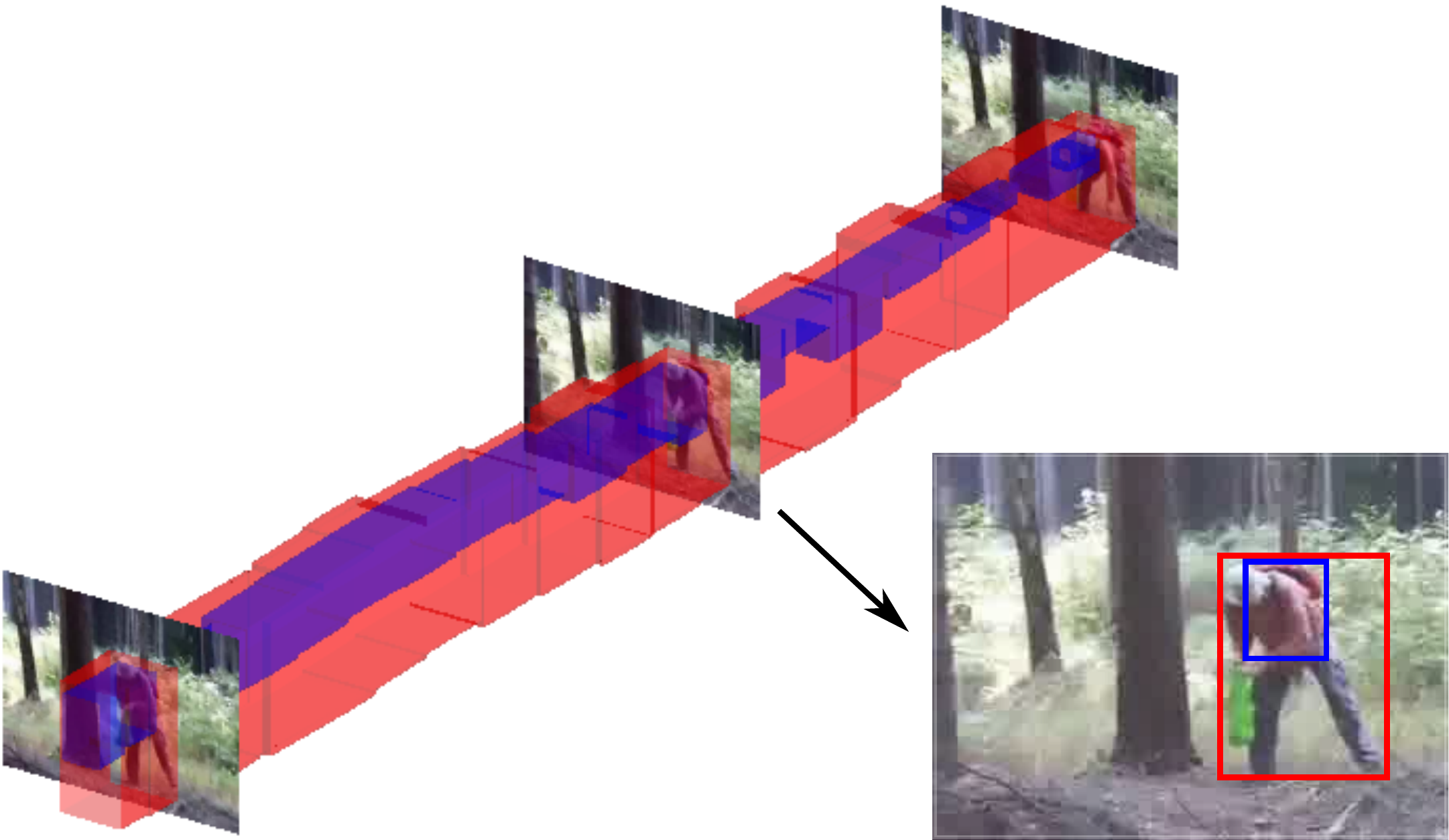}
\end{subfigure}
\hspace{0.65cm}
\begin{subfigure}{0.3\textwidth}
\centering
\emph{Sports ball (0.10) right of actor}
\includegraphics[width=\textwidth]{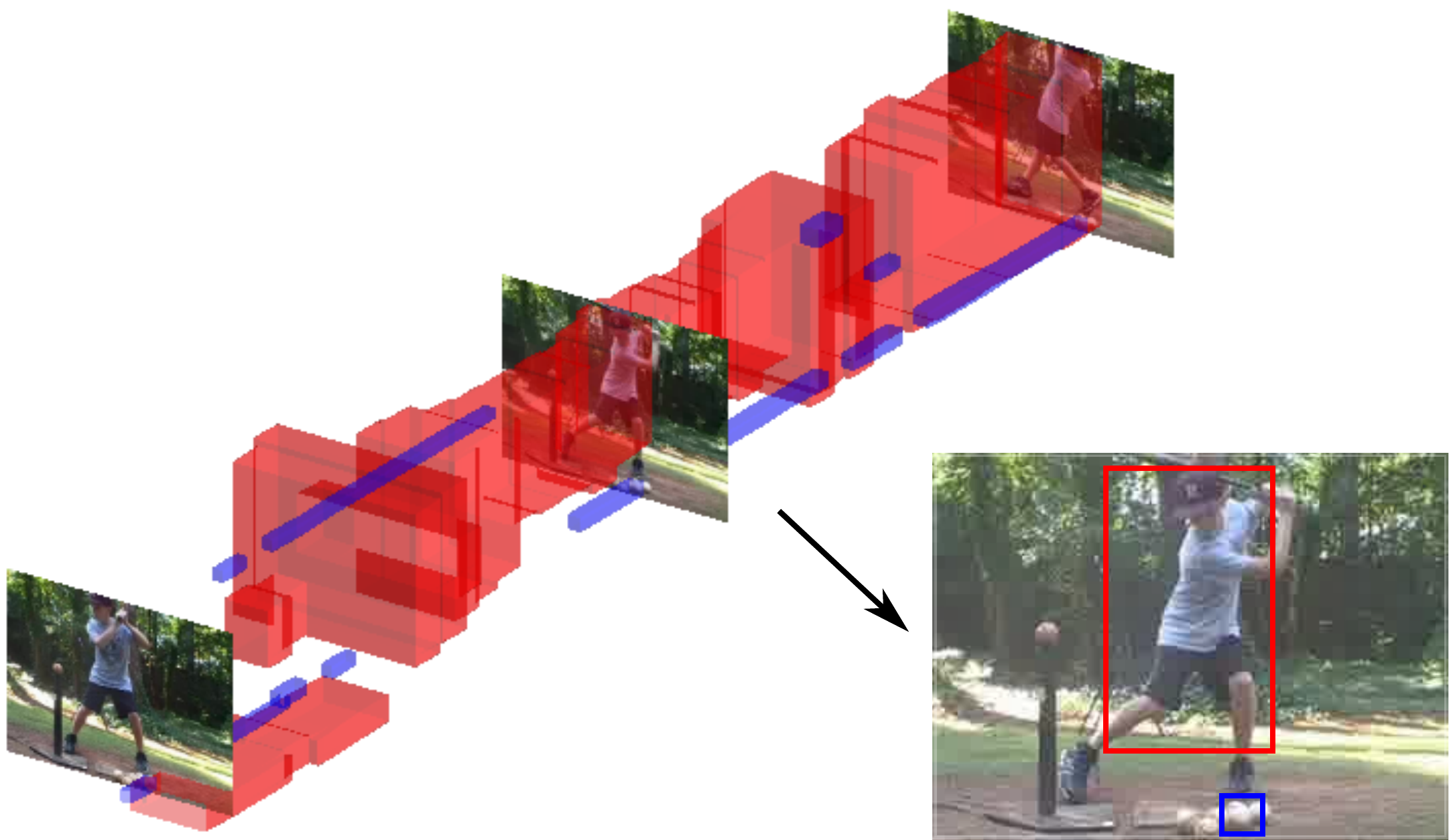}
\end{subfigure}
\hspace{0.65cm}
\begin{subfigure}{0.3\textwidth}
\centering
\emph{Sports ball (0.25) right of actor}
\includegraphics[width=\textwidth]{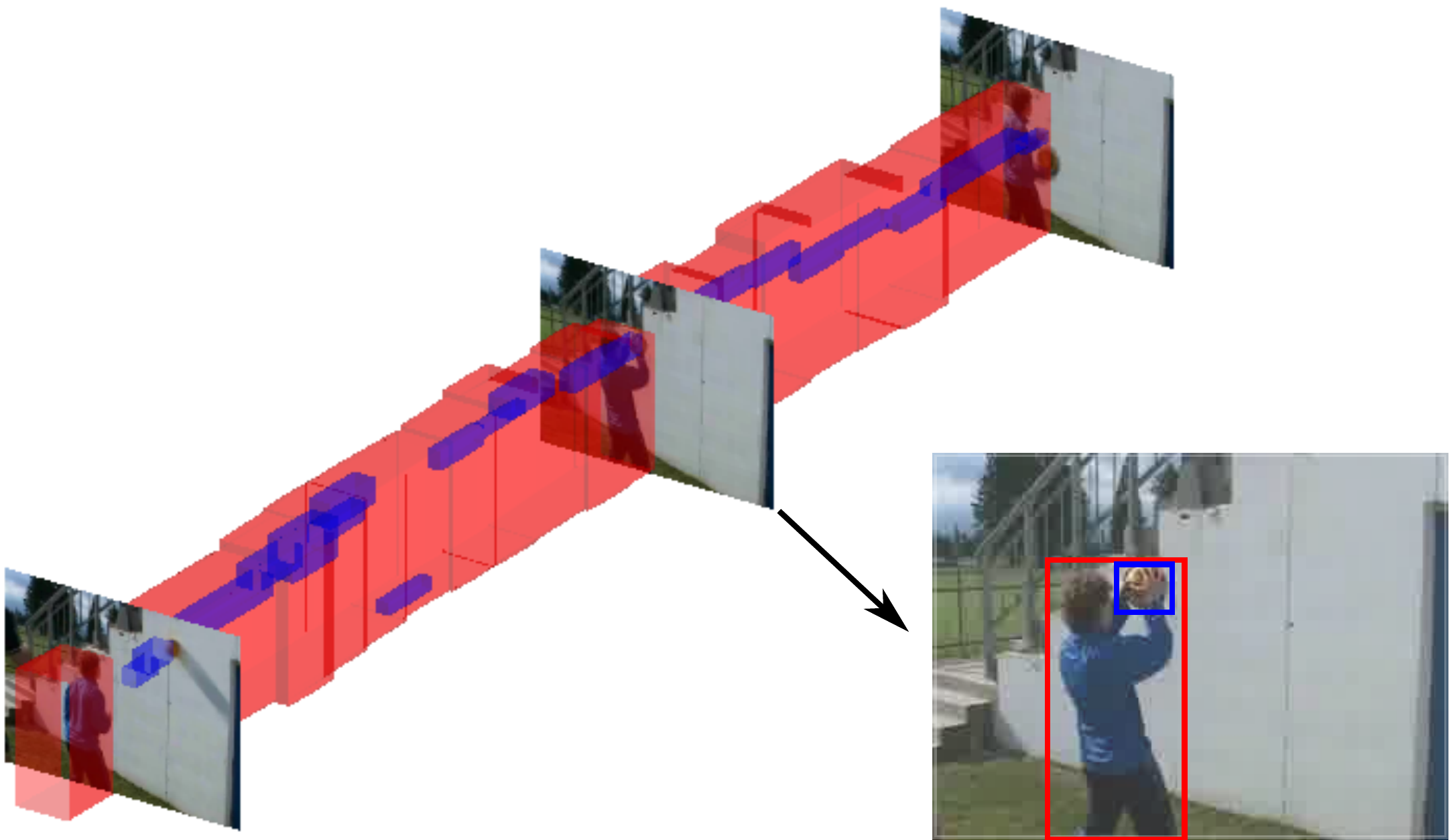}
\end{subfigure}
\caption{\textbf{Spatial-aware action retrieval.} Top retrieved results on J-HMDB given specified queries. Our retrieved localizations (red) reflect the prescribed object (blue), spatial relation, and object size.}
\label{fig:locret}
\end{figure*}

\textbf{Conclusions.}
We conclude from this experiment that our spatial-aware embedding is preferred over only using the actor and using actors and objects without spatial relations. Throughout the rest of the experiments, we will employ the spatial-aware embedding, using the top-1 object for localization and the top-5 for classification.

\subsection{Local and global object interaction}
In the second experiment, we focus on the localization and classification performance when incorporating contextual awareness from global object scores into the spatial-aware embedding.
We perform the evaluation on the UCF Sports dataset.

\textbf{Effect on localization.} In Figure~\ref{fig:exp2}, we show the AUC scores across several overlap thresholds. We show the results using our spatial-aware embedding and the combined spatial- and global-aware embedding.

We observe that across all overlap thresholds, adding global object classifier scores to our spatial-aware embedding improves the localization performance. This result indicates the importance of global object classification scores for discriminating tubes from different videos. The increase in performance is most notable at lower overlap thresholds, which we attribute to the fact that no localization information is provided by the global objects. The higher the overlap threshold, the more important selecting the right tube in each video becomes, and consequently, the less important the global object scores become.

\textbf{Effect on classification.}
In Table~\ref{tab:exp2-class}, we show the classification accuracies on the UCF Sports dataset. We first observe that our spatial-aware embedding yields results competitive to the global object approach of Jain \etal~\cite{jain2015objects2action}, who also report zero-shot classification on UCF Sports. We also observe a big leap in performance when using our spatial- and global-aware embedding, with an accuracy of 0.645. We note that the big improvement is partially due to our deep network for the global object classifiers, namely a GoogleNet trained on 13k objects~\cite{mettes2016imagenet}. We have therefore also performed an experiment with our spatial- and global-aware embedding using the network of~\cite{jain2015objects2action}. We achieved a classification accuracy of 0.374, still a considerable improvement over the accuracy of 0.264 reported in~\cite{jain2015objects2action}.

\textbf{Conclusion.} We conclude from this experiment that including global object classification scores into our spatial-aware embedding improves both the zero-shot localization and classification performance. We will use this embedding for our comparison to related zero-shot action works.

\begin{table}[t]
\centering
\begin{tabular}{l r}
\toprule
 & Accuracy\\
\midrule
Random & 0.100\\
\midrule
Jain \etal \cite{jain2015objects2action} & 0.264\\
\midrule
Spatial-aware embedding & 0.255\\
Spatial- and global-aware embedding & 0.645\\
\bottomrule
\end{tabular}
\caption{\textbf{Local and global object interaction for classification.} Adding global object awarereness improves our spatial-aware embedding considerably on UCF Sports.}
\label{tab:exp2-class}
\end{table}


\subsection{Spatial-aware action retrieval}
For the third experiment, we show qualitatively that our spatial-aware embedding is not restricted to specific action queries and spatial relations.
We show that any object, any spatial relation, and any object size can be specified as a query for spatial-aware action retrieval. For this experiment, we rely on the test videos from J-HMDB. In Figure~\ref{fig:locret}, we show three example queries and their top retrieved actions.

The example on the left shows how we can search for a specific combination of actor, object, and spatial relation.
The examples in the middle and right show that specifying different sizes for the query object leads to a different retrieval.
The examples show an interaction with a baseball (middle) and a soccer ball (right), which matches with the desired object sizes in the queries.

We conclude from this experiment that our embedding can provide spatio-temporal action retrieval results for arbitrarily specified objects, spatial relations, and object sizes.


\subsection{Comparison to state-of-the-art}
For the fourth experiment, we perform a comparative evaluation of our approach to the state-of-the-art in zero-shot action classification and localization. For localization, we also compare our results to supervised approaches, to highlight the effectiveness of our approach.

\textbf{Action classification.}
In Table~\ref{tab:exp3-class}, we provide the zero-shot classification results on the UCF-101 dataset, which provides the most comparisons to related zero-shot approaches. Many different data splits and evaluation setups have been proposed, making a direct comparison difficult. We have therefore applied our approach to the three most common types of zero-shot setups, namely using the standard supervised test splits, using 50 randomly selected actions for testing, and using 20 actions randomly for testing.

In Table~\ref{tab:exp3-class}, we first compare our approach to Jain \etal~\cite{jain2015objects2action}, who like us do not require training videos. With an accuracy of 0.328 we ouperform their approach (0.303). We also compare to approaches that require training videos for their zero-shot transfer, using author suggested splits. For the (random) 51/50 splits for training and testing, we obtain an accuracy of 0.404. Outperform the next best zero-shot approach (0.268) considerably. We like to stress that all other approaches in this regime use the videos from the training split to guide their zero-shot transfer, while we ignore these videos. When using 20 actions for testing, the difference to other zero-shot approaches increases from 0.255 \cite{kodirov2015unsupervised} and 0.311 \cite{gan2016learning} to 0.512.
The lower the number of actions compared to the number of objects in our embedding, the more beneficial for our approach.

\begin{table}[t]
\centering
\scalebox{0.92}{
\begin{tabular}{l r r r r}
\toprule
 & \textbf{Train} & \textbf{Test} & \textbf{Splits} & \textbf{Accuracy} \\
\midrule
Jain \etal \cite{jain2015objects2action} & -- & 101 & 3 & 0.303 $\pm$ 0.00\\
\rowcolor{Gray}
Ours & -- & 101 & 3 & \textbf{0.328} $\pm$ 0.00\\
\midrule
Kodirov \etal \cite{kodirov2015unsupervised} & 51 & 50 & 10 & 0.140 $\pm$ 0.02\\
Liu \etal \cite{liu2011recognizing} & 51 & 50 & 5 & 0.149 $\pm$ 0.01\\
Xu \etal \cite{xu2015semantic} & 51 & 50 & 30 & 0.186 $\pm$ 0.02\\
Xu \etal \cite{xu2017transductive} & 51 & 50 & 50 & 0.222 $\pm$ 0.03\\
Xu \etal \cite{xu2016multi} & 51 & 50 & 50 & 0.229 $\pm$ 0.03\\
Li \etal \cite{li2016recognizing} & 51 & 50 & 30 & 0.268 $\pm$ 0.04\\
\rowcolor{Gray}
Ours & -- & 50 & 10 & \textbf{0.404} $\pm$ 0.01\\
\midrule
Kodirov \etal \cite{kodirov2015unsupervised} & 81 & 20 & 10 & 0.225 $\pm$ 0.04\\
Gan \etal \cite{gan2016learning} & 81 & 20 & 10 & 0.311 $\pm$ 0.01\\
\rowcolor{Gray}
Ours & -- & 20 & 10 & \textbf{0.512} $\pm$ 0.05\\
\bottomrule
\end{tabular}
}
\caption{\textbf{Comparison to state-of-the-art} for zero-shot action classification on UCF101. For all protocols and test splits we outperform the state-of-the-art, even without us needing any training videos for action transfer.}
\label{tab:exp3-class}
\end{table}

\textbf{Action localization.}
In Table~\ref{fig:exp4-loc}, we provide the localization results on the UCF Sports, Hollywood2Tubes, and J-HMDB datasets. We first compare our result to Jain \etal~\cite{jain2015objects2action} on UCF Sports in Table~\ref{fig:exp4-loc-a}, which is the only zero-shot action localization work in the literature we are aware of. Across all overlap thresholds, we clearly outperform their approach. At the challenging overlap threshold of 0.5, we obtain an AUC score of 0.311, compared to 0.071 for Jain \etal~\cite{jain2015objects2action}; a considerable improvement.

\begin{table}[t]
\centering
\begin{subfigure}{\linewidth}
\scalebox{0.9}{
    \centering
    \begin{tabular}{l r r r r r}
    \toprule
     & \multicolumn{5}{c}{\textbf{UCF Sports}}\\
     & 0.1 & 0.2 & 0.3 & 0.4 & 0.5\\
    \midrule
    \emph{Supervised} & & & & &\\
    Gkioxari \etal \cite{gkioxari2015finding} & 0.560 & 0.560 & 0.560 & 0.520 & 0.495\\
    Jain \etal \cite{jain2014action} & 0.550 & 0.525 & 0.490 & 0.370 & 0.270\\
    Tian \etal \cite{tian2013spatiotemporal} & 0.455 & 0.425 & 0.315 & 0.265 & 0.240\\
    Cinbis \etal \cite{cinbis2014multi} & 0.292 & 0.169 & 0.128 & 0.102 & 0.049\\
    \midrule
    \emph{Zero-shot} & & & & &\\
    Jain \etal \cite{jain2015objects2action} & 0.288 & 0.232 & 0.162 & 0.099 & 0.072\\
    \rowcolor{Gray}
    Ours & \textbf{0.435} & \textbf{0.393} & \textbf{0.371} & \textbf{0.357} & \textbf{0.311}\\
    \bottomrule
    \end{tabular}
}
\caption{}
\label{fig:exp4-loc-a}
\end{subfigure}
\begin{subfigure}{\linewidth}
\centering
\scalebox{0.9}{
    \centering
    \begin{tabular}{l r r r r r}
    \toprule
     & \multicolumn{5}{c}{\textbf{Hollywood2Tubes}}\\
     & 0.1 & 0.2 & 0.3 & 0.4 & 0.5\\
    \midrule
    \emph{Supervised} & & & & &\\
    Mettes \etal \cite{mettes2016spot} & 0.345 & 0.240 & 0.154 & 0.092 & 0.048\\
    Cinbis \etal \cite{cinbis2014multi} & 0.121 & 0.051 & 0.020 & 0.007 & 0.001\\
    \midrule
    \emph{Zero-shot} & & & & &\\
    \rowcolor{Gray}
    Ours & 0.210 & 0.138 & 0.086 & 0.047 & 0.020\\
    \bottomrule
    \end{tabular}
}
\caption{}
\label{fig:exp4-loc-b}
\end{subfigure}
\begin{subfigure}{\linewidth}
\centering
\scalebox{0.9}{
    \centering
    \begin{tabular}{l r r r r r}
    \toprule
     & \multicolumn{5}{c}{\textbf{J-HMDB}}\\
     & 0.1 & 0.2 & 0.3 & 0.4 & 0.5\\
    \midrule
    \emph{Zero-shot} & & & & &\\
    \rowcolor{Gray}
    Ours & 0.346 & 0.333 & 0.305 & 0.268 & 0.230\\
    \bottomrule
    \end{tabular}
}
\caption{}
\label{fig:exp4-loc-c}
\end{subfigure}
\caption{\textbf{Comparison to state-of-the-art} for zero-shot action localization on (a) UCF Sports, (b) Hollywood2Tubes, and (c) J-HMDB. The only other zero-shot action localization approach is~\cite{jain2015objects2action}, which we outperform considerably. We also compare with several \textit{supervised} alternatives. We are competitive, especially at high overlaps thresholds.}
\label{fig:exp4-loc}
\end{table}

Given the lack of comparison for zero-shot localization, we also compare our approach to several \textit{supervised} localization approaches on UCF Sports (Table \ref{fig:exp4-loc-a}) and Hollywood2Tubes (Table \ref{fig:exp4-loc-b}). We observe that we can achieve results competitive to supervised approaches~\cite{cinbis2014multi,jain2014action,tian2013spatiotemporal}, especially at high overlaps.
Naturally, the state-of-the-art \textit{supervised} approach~\cite{gkioxari2015finding} performs better, but requires thousands of hard to obtain video tube annotations for training. Our achieved performance indicates the effectiveness of our approach, even though no training examples of action videos or bounding boxes are required. Finally, to highlight our performance across multiple datasets, we provide the first zero-shot localization results on J-HMDB in Table~\ref{fig:exp4-loc-c}.

\textbf{Conclusion.}
For classification, we outperform other zero-shot approaches across all common evaluation setups. For localization, we outperform the zero-shot localization of~\cite{jain2015objects2action}, while even being competitive to several supervised action localization alternatives.



\section{Conclusions}
We introduce a spatial-aware embedding for localizing and classifying actions without using any action video during training.
The embedding captures information from actors, relevant local objects, and their spatial relations.
The embedding further profits from contextual awareness by global objects.
Experiments show the benefit of our embeddings, resulting in state-of-the-art zero-shot action localization and classification.
Finally, we demonstrate our embedding in a new spatio-temporal action retrieval scenario with queries containing object positions and sizes.

\section*{Acknowledgements}
\vspace{-0.25cm}
This research is supported by the STW STORY project.

{\small
\bibliographystyle{ieee}
\bibliography{egbib}
}

\end{document}